%% file: anonymous-submission-latex-2024.tex
\documentclass[letterpaper]{article} 
\usepackage{aaai24}  
\usepackage{times}  
\usepackage{helvet}  
\usepackage{courier}  
\usepackage[hyphens]{url}  
\usepackage{graphicx} 

\usepackage{multirow}
\usepackage{mathrsfs}
\usepackage{enumitem}
\usepackage{bm}
\usepackage{subfigure}
\usepackage{graphics}
\usepackage{blindtext}
\usepackage{color}
\usepackage{makecell}
\usepackage{fancyhdr} 
\usepackage{algorithm}
\usepackage{algorithmic}
\usepackage{amsmath}

\usepackage{amssymb}
\usepackage{booktabs}

\allowdisplaybreaks[4]

\usepackage{fancyhdr}

\usepackage{natbib}  
\usepackage{caption} 
\usepackage{algorithm}
\usepackage{algorithmic}

\usepackage{newfloat}
\usepackage{listings}
\DeclareCaptionStyle{ruled}{labelfont=normalfont,labelsep=colon,strut=off} 
\floatstyle{ruled}
\newfloat{listing}{tb}{lst}{}
\floatname{listing}{Listing}
\title{Estimating On-road Transportation Carbon Emissions from Open Data of Road Network and Origin-destination Flow Data}
\author{
    Jinwei Zeng, Yu Liu\thanks{Corresponding author.}, Jingtao Ding, Jian Yuan, Yong Li
}
\affiliations{
    Beijing National Research Center for Information Science and Technology (BNRist),\\
    Department of Electronic Engineering, Tsinghua University, China \\
    zengjw17@gmail.com, liuyu2419@126.com
}

\usepackage{bibentry}

\begin{document}

\maketitle

\begin{abstract}
Accounting for over 20\% of the total carbon emissions, the precise estimation of on-road transportation carbon emissions is crucial for carbon emission monitoring and efficient mitigation policy formulation. However, existing estimation methods typically depend on hard-to-collect individual statistics of vehicle miles traveled to calculate emissions, thereby suffering from high data collection difficulty. To relieve this issue by utilizing the strong pattern recognition of artificial intelligence, we incorporate two sources of open data representative of the transportation demand and capacity factors, the origin-destination (OD) flow data and the road network data, to build a \underline{H}ierarchical h\underline{E}terogeneous graph learning method for o\underline{N}-road \underline{C}arbon \underline{E}mission estimation (\textbf{HENCE}). Specifically, a hierarchical graph consisting of the road network level, community level, and region level is constructed to model the multi-scale road network-based connectivity and travel connection between spatial areas. Heterogeneous graphs consisting of OD links and spatial links are further built at both the community level and region level to capture the intrinsic interactions between travel demand and road network accessibility. Extensive experiments on two large-scale real-world datasets demonstrate HENCE's effectiveness and superiority with $R^2$ exceeding 0.75 and outperforming baselines by 9.60\% on average, validating its success in pioneering the use of artificial intelligence to empower carbon emission management and sustainability development. The implementation codes are available at this link: \url{https://github.com/tsinghua-fib-lab/HENCE}.
\end{abstract}

\input{1.intro}

\input{2.related}
\input{4.method}

\input{5.experiments}

\input{7.conclusion}

\bibliography{reference}

\end{document}

%% file: 1.intro.tex
\section{Introduction}
Alarmed by the global warming trend, countries are making joint efforts to mitigate carbon emissions and achieve sustainable development~\cite{unfccc, lashof1990relative}. On-road transportation carbon emissions, taking up a large share of 28\% of the total carbon emissions in the United States~\cite{citaristi2022international}, is the crucial and prioritized carbon mitigation target~\cite{song2012greenhouse, wang2015carbon}. Thereby, it is essential to develop techniques to precisely estimate on-road carbon emissions since the comprehension of emission distribution and magnitude is fundamental for formulating targeted and efficient carbon mitigation policies~\cite{zhang2019review}.

\begin{figure}[t]
    \centering
    \includegraphics[width=0.85\columnwidth]{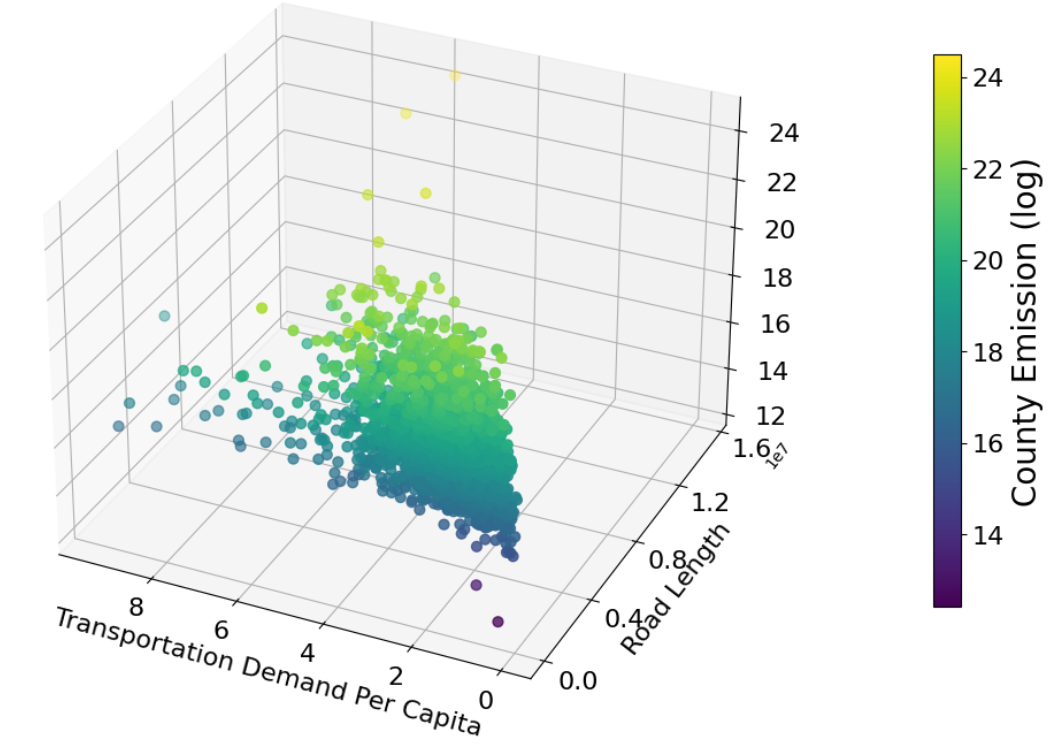}
    \caption{Visualization of road network length and per capita transportation demand for counties in the United States. Here the color corresponds to the county's on-road carbon emissions. Data sources are consistent with the experimental setting.}
    \label{fig:visual}
\end{figure}

On-road carbon emissions are primarily affected by travel demands and road network conditions~\cite{gomez2009driving}: While travel demands generally determine the approximate travel distance level, the road network connectivity and layout impact the routing and traffic situations. Existing on-road carbon emission estimation methods typically rely on collected travel miles statistics to learn how the region's travel demand is satisfied by the road network's connectivity, thereby calculating the corresponding carbon emission. Specifically, these emission estimation methods collect mileage data from either the roadway perspective or the vehicle perspective. Roadway-based methods monitor flows on each road to calculate the corresponding total vehicle miles traveled, while vehicle-based methods aggregate individual vehicle trajectory statistics to obtain the total miles~\cite{gately2019darte, gately2015cities, perugu2019emission, hui2007evaluation}. The collected mileage data are then multiplied by the empirical emission per unit mileage statistic to calculate the on-road carbon emissions~\cite{liu2015reduced}. Since both types of methods collect data with a bottom-up paradigm where the coverage and precision of the collected individual data are highly required, they all suffer from high data collection efforts and expenses.

The recent boom of artificial intelligence methods, which have shown their strengths in automating the process of pattern mining and indicator prediction~\cite{aiken2022machine, wang2016crime, dong2019predicting}, and the availability of various sources of open data, present a promising solution for on-road carbon emission estimation with low data collection cost: On one hand, a trained prediction model can spare the high data collection expenses inherent in existing carbon estimation methods for prediction targets. On the other hand, with the investigated correlations between on-road carbon emission and its impact factors, we can better estimate on-road carbon emission changes under factor interventions, thereby gaining valuable insights for efficient carbon-mitigation policy formulation.

Therefore, corresponding to the two primary factors of on-road carbon emissions, traveling demand, and road network conditions, we introduce two open data sources representative of the factors, the origin-destination (OD) flow data and road network data, and build a graph learning model to predict on-road carbon emissions with them. While the transportation demands determine the general travel distance levels, the varying road network connectivity across the regions leads to distinct transportation efficiency, thereby impacting the on-road carbon emissions. The scatterplot shown in Figure~\ref{fig:visual} also proves evidence for the fact that county carbon emissions are highly correlated with transportation demand per capita and road length, validating the sufficient inclusion of on-road carbon emission information in OD data and road network data. However, comprehensively modeling the joint impacts of road networks and OD flows on on-road carbon emissions faces several challenges, which can be formulated into the following two aspects:

\begin{itemize}
[leftmargin=5pt]
    \item \textbf{Road network and OD network interact at multiple spatial scales of the region where different emission patterns exist.} While long-distance travel usually relies more on motorways that connect spaces at the region scale, short-distance travel is mostly fulfilled by regular roads within zones. The driving patterns vary on different types of roads, and so do the corresponding emissions. It is essential to explicitly model how such multi-scale transportation is fulfilled by the hierarchical road network system to estimate the on-road carbon emissions accurately.
    \item \textbf{Road networks and ODs impact on-road carbon emissions in a coupled way with complex and heterogeneous interactions.} A region's on-road carbon emission is the matching consequence of road network accessibility and transportation demand across the region, where both factors vary a lot across space, and so is their interaction.
\end{itemize}

To address the aforementioned challenges, we construct our \underline{H}ierarchical h\underline{E}terogeneous graph learning model for o\underline{N}-road \underline{C}arbon \underline{E}mission estimation (\text{HENCE}). Considering the multi-scale nature of connectivity and travel relations between areas, we construct a hierarchical graph comprised of the road network level, community level, and region level to model the relations at various scales explicitly. While the community-level graph captures between-community road network connectivity for short-distance intra-region OD modeling, the region-level graph is designed to model the matching degree between inter-region OD flows and region-level connectivity represented by motorways. To further capture the intricate interactions between road network connectivity and transportation demand, we construct a heterogeneous graph for both the community level and the region level, where two types of edges corresponding to the spatial road network connectivity and travel relations respectively are built and attentional message aggregation is developed to model their interactions. With the bottom-up pooling and message-passing mechanism, we obtain the representations of the regions fully characterizing how transportation demands are met by their road networks and predict on-road carbon emissions with the representations. Overall, the main contribution of this work can be summarized into the following aspects:

\begin{itemize}
[leftmargin=5pt]
    \item To address the severe issue of high data collection difficulty in existing on-road carbon emission estimation methods that hinders efficient carbon emission estimations, we develop a prediction model that incorporates open data to estimate on-road transportation carbon emissions. Two sources of open data indicating the regional transportation demand and spatial connectivity respectively are incorporated: OD flow data and road network data. 
    \item We propose a hierarchical heterogeneous graph learning model that comprehensively models how diverse transportation demands are satisfied by road network-based spatial connectivity at various scales. 
    \item Experimental results on two large-scale real-world datasets validate our model's effectiveness, with $R^2$ exceeding 0.75 and outperforming existing state-of-the-art methods by 9.60\%. The outstanding performance of our model also demonstrates the potential of artificial intelligence in contributing to sustainable development. 
\end{itemize}

%% file: 2.related.tex
\section{Related Works}
\textbf{Traditional On-road Carbon Emission Calculation Methods.} 
Traditional carbon emission estimation methods typically rely on multiplying the total traveling distances of various vehicle types by empirical carbon emission amount per unit mileage. Therefore, the key point lies in the acquisition of travel distances for different modes of transportation. Existing inventories and models typically collect bottom-up travel distances from the roadway perspective or the vehicle perspective. Specifically, DARTE~\cite{gately2019darte, gately2015cities}, the first annual on-road carbon emissions database for the conterminous USA, relies on Federal Highway Administration's Highway Performance Monitoring System to monitor the roadway-level vehicle miles traveled. MOVES~\cite{perugu2019emission}, as a representative of mobile source emission models~\cite{hui2007evaluation, liu2020variation}, takes individual vehicle travelling information, such as trajectories and speeds, as inputs to accurately estimate vehicle emissions. However, both perspectives have a high requirement for data collection infrastructure, resulting in heavy collection expenses and difficulty in large-scale estimation across time and space.   

\hspace{\fill} \\
\noindent
\textbf{Data-driven On-road Carbon Emission Estimation Methods.} With the advancements of machine learning and the availability of tremendous data, several data-driven methods have been employed to estimate carbon emissions efficiently. Concerning road-level emission estimation, a three-layer perception neural network was built to utilize the characteristics of collected vehicle and road network data to infer road-level on-road carbon emissions~\cite{lu2017predicting}. About regional on-road carbon emission prediction, several works~\cite{khajavi2023predicting, huo2023prediction} introduce machine learning methods, including SVM and random forest, to explore the impacts of road conditions, travel demand, and other factors on on-road carbon emissions. However, these methods rely on the correlation between aggregated regional statistics and regional on-road carbon emissions, failing to model the heterogeneous spatial distribution of road networks and travel demands that may exist for regions with similar aggregated statistics.

%% file: 4.method.tex
\section{Method}
\begin{figure}[t]
    \centering
    \includegraphics[width=1.\columnwidth]{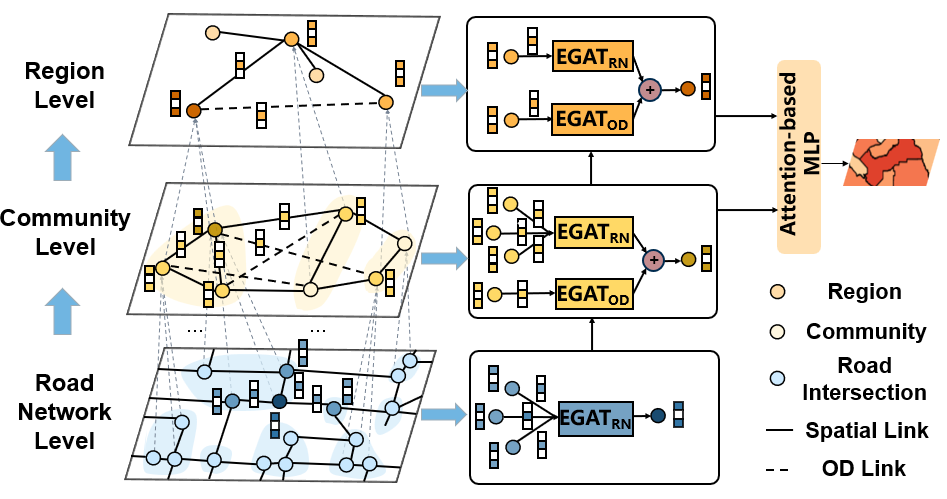}
    \caption{Our hierarchical heterogeneous graph learning method for on-road carbon emission estimation.}
    \label{fig:multilayer}
\end{figure}

\subsection{Model Framework}
To leverage the capacity of artificial intelligence for modeling the joint impact of the road network and origin-destination flow on on-road carbon emissions, we propose the ~\text{HENCE} model. First, a hierarchical graph is constructed to characterize the multi-scale road network connectivity relations and travel relations between areas, in which heterogeneous graphs are incorporated at several levels to model the interactions between road network-based spatial accessibility and travel demand. Furthermore, a multi-level information fusion and prediction module is designed to characterize the differences in regions' travel outwardnesses.  

The construction scheme of the hierarchical heterogeneous graph is visualized in Figure~\ref{fig:multilayer}, which consists of three levels: the road network level, the community level, and the region level. Different levels are connected based on the affiliation relationships between them. At the road network level, a graph is constructed according to the topology of the road network, and attributes of road intersections and segments are modeled as initial features of the nodes and edges respectively. To effectively abstract local road network structures and model how intra-region short-distance travel are satisfied by the road network connectivity inside the region, we construct a community-level heterogeneous graph for each region with community nodes and two types of edges. Specifically, the community node describes the intra-community road network conditions whose features are pooled from the road network level. For the two types of edges, the spatial link represents the road network-based connectivity between adjacent areas, and the OD link describes the travel demand between areas. However, considering the indispensable amount of cross-region transportation that associates areas at the region scale, we further construct a region-level heterogeneous graph to characterize how inter-region transportation demands are met by between-region road network connectivity. In this way, we obtain the intra-region embeddings by aggregating the community-level representations and the inter-region embeddings by message passing through the region-level graph. An attention-based multi-layer perceptron (MLP) adaptively aggregates these two embeddings and makes the final estimations of on-road carbon emissions. 

In the following sections, we will first give a detailed elaboration on the heterogeneous graph learning module that is used at both the community level and the region level. Then more details will be presented on the information propagation of the hierarchical graph that models the multi-scale connectivity relations and travel relations between areas. Furthermore, we will present how to fuse the learned representations for final on-road carbon emission estimations.

\subsection{Heterogeneous Graph Learning Module for Road Network and OD Interaction Modeling}
Travel demand and road network connectivity jointly impact carbon emissions of on-road vehicles as evidenced by Figure~\ref{fig:visual}. While the travel demand generally determines the vehicle travel distances, the road network structures and conditions affect the specific routing and travel speeds of each journey, thereby further influencing carbon emissions.

Therefore, we construct a heterogeneous graph whose two types of edges correspond to such two types of factors: spatial links that are built between spatially adjacent areas to characterize road network connectivity and OD links that are built between areas to represent mutual transportation demand. For the node of the graphs, its features are the representation of transportation information within the node's semantic area that is initially pooled from the lower level. The features of the spatial links are the pooled road network representations for roads connecting these two nodes, and the features of the OD links are the origin-destination flow values for OD pairs. 

Based on the constructed heterogeneous graph, we further develop the message-passing mechanism to model the intricate interactions between road networks and travel demand. Considering the edge-featured graph attentional (EGAT) convolutional layer's capability to utilize edge features and node features jointly, we adopt it as the basic message-passing method to propagate information for each type of edge~\cite{wang2021egat}. Denoting the node features as $\bm{V}$ and the edge features as $\bm{E}$, the EGAT layer propagates information as formulated:

\begin{equation}
{\bm{V}_{i}^{'}} = \sum_{j\in \mathcal{N}_i}\alpha_{ij}\cdot \bm{W}\bm{V}_{j},
\end{equation}

\begin{equation}
\bm{E}_{ij}^{'} = \bm{A}[\bm{V}_{i}||\bm{E}_{ij}||\bm{V}_{j}].
\end{equation}
where 
\begin{equation}
\alpha_{ij} = \text{Softmax}\big(h_{ij}\big),
\end{equation}

\begin{equation}
h_{ij} = \text{LeakyReLU}\big(\mathbf{a}^{\top}\bm{U}[\bm{V}_{i}||\bm{E}_{ ij}||\bm{V}_{j}]\big),
\end{equation}
Here $\bm{V}_{i}^{'}$ is the updated node representation for node $i$, $\bm{E}_{ij}^{'}$ is the updated edge representation for edge $ij$. $\mathcal{N}_i$ refers to node $i$'s one-hop neighborhood, $\alpha_{ij}$ is the attention score between node $i$ and node $j$, $\mathbf{a}$ is the attentional weight vector, and $\bm{W}$, $\bm{A}$ and $\bm{U}$ are the learnable weight matrices.

Therefore, the road network connectivity of a region is represented by the information propagation through spatial links ($rn$) and the transportation demand is represented by the information propagation through OD links ($od$): 

\begin{equation}
\bm{V}_{m}^{'}, \bm{E}_{m}^{'} = \text{EGAT}\big(\bm{V}, \bm{E}_{m}\big), m \in \big\{rn, od\big\},
\label{equ:egat}
\end{equation}

An attentional aggregation follows to characterize the interactions between road network and transportation demand:  
\begin{equation}
   \bm{V}^{'} = \sum_{m \in \big\{rn, od\big\}} \beta_{m}\cdot {\bm{V}^{'}_{m}},
   \label{equ:agg}
\end{equation}
where
\begin{equation}
   \alpha_{m} = \bm{c}^{T}\cdot \text{Tanh}\big(\bm{W}\cdot {\bm{V}^{'}_{m}} + b\big), m \in \big\{rn, od\big\},
   \label{equ:attention}
\end{equation}

\begin{equation}
   {\beta_{m}} = \frac{\text{exp}\big({\alpha_{m}}\big)}{\sum_{i \in \big\{rn, od\big\}} \text{exp}\big({\alpha_{i}}\big)}.
\end{equation}

By repeating the calculation process in Equ~\ref{equ:egat} and Equ~\ref{equ:agg} for $L$ times where $L$ is a tunable parameter, we obtain the heterogeneous graph node representations ${\bm{V}^{'}}$ that fully characterize the joint impacts of road network connectivity and transportation demand. 

\subsection{Hierarchical Graph Learning Module for Multi-scale Information Propagation}
The hierarchical graph has two essential roles, which are to hierarchical connectivity inherent in road networks and to comprehensively consider the impacts of multi-scale transportation demands between areas on on-road carbon emissions. In this section, we will present how these two roles are accomplished by the information propagation and aggregation mechanism of the three-level hierarchical graph.  

As on-road carbon emissions occur on the road networks, at the bottom level, we construct a road network graph where nodes correspond to road intersections and edges correspond to road segments. To fully leverage the road intersection attributes (relative longitude and latitude of the intersection in the county and degree) and road segment attributes (relative longitude and latitude of the road centre, road length, and road class) to learn the connectivity information of these road network elements, we also employ the EGAT convolutional layer~\cite{wang2021egat} to propagate information on this graph. Denoting road intersection attributes as $\bm{V}_r$ and road segment attributes as $\bm{E}_r$, this process can be formulated as:

\begin{equation}
\bm{V}^{'}_r, \bm{E}^{'}_r = \underbrace{\text{EGAT}\big(...\big(\text{EGAT}}_{L_r\  \text{layers}}\big(\bm{V}_r, \bm{E}_r\big)\big)\big).
\end{equation}

To obtain the spatial connectivity information for different parts of the region and model how the spatial connectivity satisfies transportation demands, we construct a community-level heterogeneous graph with both spatial links and OD links for each region. We pool information for road intersections and road segments inside each community as the initial community attributes characterizing its inner road network structure and connectivity. For communities with spatial links, we pool the representations of road segments connecting these communities as spatial link attributes, thereby characterizing the road network connectivity between communities. Meanwhile, we input the OD flow values between communities to serve as the OD link attributes. The pooling process can be formulated as follows:

\begin{align}
&\bm{V}_{c, m} = \text{Concat}\big(\big[\phi\big(\{\bm{V}^{'}_{r,i}\}|_{i\in {\mathcal{C}}_{m}} \big), \phi\big(\{\bm{E}^{'}_{r,ij}\}|_{i, j\in {\mathcal{C}}_{m}}\big)\big]\big),\\
&\bm{E}_{c, mn, rn} = \phi\big(\{\bm{E}^{'}_{r,ij}\}|_{i\in {\mathcal{C}}_{m},j\in {\mathcal{C}}_{n}} \big), 
\end{align}
where $\phi$ is the optional pooling function among sum, max, mean pooling, etc. ${\mathcal{C}}$ is short for communities.

With the built community-level graph, we adopt the heterogeneous graph learning module to model how the road networks meet the intra-region transportation demands. By pooling its outputs, the community representations $\bm{V}^{'}_c$, we obtain the region-level intra-region representations $\bm{V}_{g, intra}$. 

Considering the vast amount of cross-region transportation and their indispensable contribution to regional on-road carbon emissions as well, we build a region-level heterogeneous graph to model the interaction between region road network connectivity and cross-region transportation demands. The construction of the region-level heterogeneous graph is identical to the community-level ones, whose node features are the intra-region representations pooled from the community level, and the spatial link and OD link correspond to spatially adjacent regions and OD-connected regions respectively. By applying the heterogeneous graph learning module to this graph, we comprehensively model the interactions between regions and obtain the inter-region representations $\bm{V}_{g, inter}$. To mention, as calculating one region's inter-region representation involves the pooled representations of neighboring regions, it will thereby involve extensive computations on these regions' community-level and road network-level graphs. To control the computational cost of our model, the region attributes are only updated after a complete training cycle, which is one epoch. The process can therefore be formulated as: For the training epoch $T$, the inter-region representations ${{\bm{V}_{g, inter}}}^{(T)}$ are calculated with ${{\bm{V}_{g, intra}}}^{(T-1)}$ which are the intra-region representations generated by the model trained after $T-1$ epochs. In this way, we balance the computational costs and model training efficiency.

\subsection{Multi-scale Information Fusion and Prediction}
As regions have different traveling outwardness, the contributions of intra-region ODs and inter-region ODs on on-road carbon emissions vary across regions. Therefore, we use the attention fusion functions in Equ~\ref{equ:agg} to aggregate intra-region representations derived from the community scale and the inter-region representations derived from the region scale into the ultimate region representations ${\bm{V}_{g}}$.

An MLP model is introduced to predict the on-road carbon emission for region $i$. Denoting the prediction results as $\{\hat{Y_i} \}$, we adopt the mean squared error loss function for loss calculation and parameter optimization for region set $\mathcal{A}$:

\begin{equation}
   \mathcal{L} = \frac{1}{|\mathcal{A}|}\sum_{i \in \mathcal{A}}{\big({Y_i} - {\hat{Y_i}}\big)^2}.
   \label{equ:loss}
\end{equation}

%% file: 5.experiments.tex
\section{Experiment}


\subsection{Experimental Setups}
\subsubsection{Dataset}
Due to the scarcity of reliable on-road carbon emission databases, we conduct experiments on two prediction years of the well-acknowledged American on-road carbon emission database, DARTE~\cite{gately2015cities, gurney2009high}\footnote{\url{https://daac.ornl.gov/CMS/guides/CMS_DARTE_V2.html}}. DARTE relies on a national monitoring system to monitor the vehicle counts and total miles traveled on US roads, based on which on-road carbon emission statistics are calculated for all counties in the United States. Since its latest released year is 2017, we set our experimental tasks as predicting regional on-road carbon emissions for 2015 and 2017 between which ODs and road networks of the United States have developed to a certain extent. For each year, we obtain its corresponding road network data from the OpenStreetMap\footnote{\url{https://planet.openstreetmap.org/planet/full-history/}}, and OD data from Longitudinal Employer-Household Dynamics program\footnote{\url{https://lehd.ces.census.gov/data/#lodes}}, which publishes fine-grained commuting ODs based on employment statistics obtained by the tax authority. According to the administrative subdivisions of the United States, the county corresponds to the region and the census tract corresponds to the community in our experiment.


\subsubsection{Baselines}

\textbf{Carbon Prediction Machine Learning Methods:} (1) \textbf{SVM}~\cite{huo2023prediction} Support Vector Machine, which finds a hyperplane that distinctly regresses the data points. (2) \textbf{Random Forest}~\cite{khajavi2023predicting}. A decision tree method using bagging ensemble strategy.

\begin{table*}[t]
    \centering
    \begin{tabular}{c c c c c c c c}
        \toprule 
        & &  \multicolumn{3}{c}{Year 2015} &  
        \multicolumn{3}{c}{Year 2017}\\
        \cmidrule{3-5} \cmidrule{6-8} 
        Groups & Models &  $R^2$ & MAE & RMSE  & $R^2$ & MAE & RMSE\\
        \midrule
        \multirow{2}*{\makecell[c]{Carbon Prediction Methods}} & SVM &0.5083&0.7003&0.9361&0.5702&0.6767&0.9222 \\
        ~ & Random Forest &\underline{0.6708}&\underline{0.5892}&\underline{0.7659}&\underline{0.7325}&\underline{0.5148}&\underline{0.7276} \\
        \cmidrule{1-8}
        \multirow{4}*{\makecell[c]{Graph Learning Methods}} & EGAT &0.5386&0.6628&0.9068&0.6786&0.5685&0.7990\\
        ~ & EGNN&0.5296&0.6447&0.9157&0.6324&0.6126&0.8530\\
        ~ & Sortpooling&0.3287&0.8254&1.0940&0.3988&0.8351&1.0910 \\
        ~ & DiffPool &0.3498&0.8002 &1.0767&0.4751&0.7340&1.0191 \\
        \cmidrule{1-8}
        \multirow{1}*{\makecell[c]{Road Representation Method}}& RFN &0.3091&0.8491&1.1099&0.4570&0.7559&1.0367\\
        \midrule
        \multirow{2}*{\makecell[c]{}} & \textbf{HENCE} & $\bm{0.7502}$& $\bm{0.4890}$ & $\bm{0.6672}$ &$\bm{0.7865}$ & $\bm{0.4541}$ & $\bm{0.6500}$  \\
        ~&\textbf{Improv.}  &$\bm{11.83\%}$&$\bm{17.01\%}$ &$\bm{12.89\%}$&$\bm{7.37\%}$&$\bm{11.79\%}$&$\bm{10.67\%}$ \\
        \bottomrule
    \end{tabular}
    \caption{Overall performance of HENCE in two datasets. Bold denotes best results and underline denotes second-best ones.}~\label{tbl:results}
\end{table*}

\noindent

\noindent\textbf{Graph Learning Methods:}
We compare our HENCE model with methods incorporating new message-passing or pooling mechanisms. (1) \textbf{EGAT}~\cite{wang2021egat}. The edge-featured graph attention network integrates edge features in the calculation of the message and attention weights. (2) \textbf{EGNN}~\cite{satorras2021n}. The equivariant graph neural network represents graphs in a way equivariant to rotations, translations, reflections and permutations. (3) \textbf{SortPooling}~\cite{zhang2018end}. A pooling mechanism that sorts the nodes and selects a subset of top-ranked nodes for pooling. (4) \textbf{DiffPool}~\cite{ying2018hierarchical}. A differentiable graph pooling module to learn hierarchical pooled representations of graphs.

\noindent\textbf{Road Representation Learning Methods:} (1) \textbf{RFN}~\cite{jepsen2019graph}. This work proposes a relational fusion network to aggregate node relations and edge relations to represent the road network.  

For fair comparisons, all baselines are inputted with road network and OD information. For machine learning methods, we construct the input attributes combining road network features (including the total number of intersections, roads, and the total road length), and aggregated OD information (including total inflow, outflow, count of intra-county OD, and count of inter-county OD). For graph learning and road representation methods, we concatenate OD information attributes with the learned road network representation vectors for final prediction. Since EGAT~\cite{wang2021egat}, EGNN~\cite{satorras2021n}, and RFN~\cite{jepsen2019graph} only output representations at the node and edge level, we use a mean pooling layer to pool the learned node and edge representations to the region level for region-level on-road carbon emission estimation.  

\subsubsection{Metrics and Implementation}
Three evaluation metrics commonly used for evaluating indicator prediction are selected: coefficient of determination ($R^2$), mean absolute error (MAE), and rooted mean squared error (RMSE)~\cite{han2020lightweight, jean2016combining, xi2022beyond}. We apply Adam optimizer for parameter learning and perform a grid search on all hyperparameters of both our model and baselines, with a search range of the number of graph convolutional layers in \{2, 3, 4\}, learning rate in [1e-4, 5e-2], and batch size in \{8, 16, 32, 64\}. We adopt the mean pooling layer as the pooling function in our model's implementation.

\subsection{Overall Performance}
Comparisons on Year 2015 and Year 2017 dataset presents us with the following findings:
\begin{itemize}
[leftmargin=5pt]
    \item \textbf{Our model significantly outperforms existing state-of-the-art methods}. As shown in the dataset, compared with the best baseline method, our model improves $R^2$  by 9.60\% and reduces MAE by 14.40\% on average. The consistent and significant improvement of our method on all metrics validates the effectiveness of our method in capturing how the complex interactions between road networks and OD impact on-road carbon emissions.  
    \item \textbf{Our model is capable of capturing the intrinsic correlations between OD, road networks, and on-road carbon emissions.} For 2015 and 2017 between which road network and OD have both evolved, our model all achieves an outstanding improvement over baselines. Therefore, we can conclude that our model is capable of capturing diverse correlations between OD patterns and road networks across different datasets.
    \item \textbf{Existing methods fail to capture the multi-scale intrinsic relations between road network and OD network.} Feature-based carbon prediction machine learning methods ignore the spatiality nature of road networks and OD networks. While EGAT~\cite{wang2021egat} and EGNN~\cite{satorras2021n} leverage graphs to represent road networks, the flat graph design makes it hard to capture the multi-scale relations between areas. Meanwhile, the complex heterogeneity of local road network structures contributes to the difficulty of adaptively abstracting road networks, which may explain the poor performance of DiffPool~\cite{ying2018hierarchical}. Moreover, existing graph learning models lack mechanisms to model the interactions between road networks and transportation demand, thereby failing to capture their joint impacts on on-road carbon emissions. 
\end{itemize}

\subsection{Ablation Study}
Our proposed HENCE model consists of three key components: the hierarchical graph abstracting the complex road network, and two heterogeneous graphs at both the community and region level modeling the interactions between road network connectivity and intra-region and inter-region transportation respectively. To evaluate the contribution of the hierarchical graph structure, we conduct experiments on two variants: Our model excluding the community-level modeling and the region-level modeling respectively. To validate the necessity of constructing the heterogeneous graph that adaptively models the interactions between road network-based spatial connectivity and transportation demands, we exclude the spatial link and OD link respectively from all heterogeneous graphs of our model.
\begin{table}[t]
    \centering
    \begin{tabular}{c c c c c c}
        \toprule 
        & \multicolumn{2}{c}{Year 2015} &&  
        \multicolumn{2}{c}{Year 2017}\\
        \cmidrule{2-3} \cmidrule{5-6} 
        Models &$R^2$ & MAE  && $R^2$ & MAE\\
        \midrule
        w/o spatial link &0.652&0.578 &&0.719&0.533 \\
        w/o OD link &0.689&0.610 &&0.706&0.560 \\
        w/o CL &0.722&0.531 &&0.761&0.494\\
        w/o RL &0.601&0.610 &&0.668&0.571 \\
        \textbf{HENCE} &\textbf{0.750}&\textbf{0.489} &&\textbf{0.787}&\textbf{0.454} \\
        \bottomrule
    \end{tabular}
    \caption{Ablation studies on two datasets. Here CL is short for community level and RL is short for region level.}
    \label{tbl:ablation}
\end{table}

The ablation evaluation results are reported in Table~\ref{tbl:ablation}. We can observe that excluding either spatial links or OD links from our model's community level and region level reduces the performance substantially, leading to a 10.84\% and 9.22\% decrease of $R^2$ on average. Therefore, it can be concluded that spatial connectivity information and transportation demand information are vital in estimating on-road carbon emissions, emphasizing the critical role of the heterogeneous graph learning module in comprehensively modeling their interactions. We further compare the performance of our HENCE model with the variants losing either the community-level graph or the region-level graph. Losing either level of the hierarchical graph leads to a substantial performance decrease: Removing the community-level modeling brings a 3.51\% decrease of $R^2$ and an 8.65\% decrease of MAE on average. This validates the effectiveness of the community-level graph for abstracting diverse local road network structures and modeling the interaction between intra-region ODs and the road network. Moreover, removing the region-level graph decreases $R^2$ by 17.46\% on average. Such a substantial decrease indicates the higher contribution of inter-region transportation to on-road carbon emissions, thereby emphasizing the importance of modeling how road network conditions meet inter-region transportation demands.

\subsection{Transferability Study}
\subsubsection{Spatial Transfer Study}
Through the experiments above, we have comprehensively validated the effectiveness of our model. However, in real applications, we may face label insufficiency problems where we lack enough on-road carbon emission training labels for our prediction targets. Therefore, it is of practical importance to investigate our model's spatial transferability. We divided the counties of the United States into two sets of source-target pairs based on an east-west and north-south division\footnote{https://github.com/tsinghua-fib-lab/HENCE} and conducted a transferability test on the Year 2017 dataset. 

As shown in Table~\ref{tbl:transfer}, HENCE achieves the best experimental results in both transfer scenarios. The average performance gain of HENCE over the baseline with the best performance, Random Forest~\cite{khajavi2023predicting}, is 13.43\% on $R^2$ metric and 13.81\% on MAE. The high values of $R^2$ performances further validate our model's capacity of handling practical transfer applications. Therefore, we can conclude that our model can capture the complex intrinsic correlations between road network, OD network, and on-road carbon emissions that generalize across regions.  
\begin{table}[t]
    \begin{tabular}{c c c c c}
        \toprule 
        &  \multicolumn{2}{c}{West-east} &  
        \multicolumn{2}{c}{North-south}\\
        \cmidrule{2-3} \cmidrule{4-5} 
        Models &  $R^2$ & MAE & $R^2$ & MAE \\
        \midrule
        SVM &0.6067&0.6415&0.5133&0.6818\\
        RF &\underline{0.7008}&\underline{0.5155}&\underline{0.6298}&\underline{0.5625}\\
        EGAT&0.5422&0.6558&0.5696&0.5917\\ EGNN&0.6071&0.6510&0.5801&0.5889\\
        SortPooling &0.3978&0.8373&0.3383&0.8776 \\
        DiffPool &0.4247&0.8134&0.3927&0.8057\\
        RFN &0.4220&0.8170&0.3665&0.8369\\
        \textbf{HENCE} & $\bm{0.7811}$& $\bm{0.4487}$ &$\bm{0.7267}$& $\bm{0.4801}$ \\
        \cmidrule{1-5}
        Improv. &$\bm{11.46\%}$&$\bm{12.96\%}$ &$\bm{15.39\%}$&$\bm{14.65\%}$\\
        \bottomrule
    \end{tabular}
    \centering
    \caption{Experimental results of HENCE and existing state-of-the-art methods in two transfer scenarios.}~\label{tbl:transfer}
\end{table}

\subsubsection{Temporal Transfer Study}
\begin{table}[t]
    \begin{tabular}{c c c c}
        \toprule 
        Models &$R^2$ & MAE & RMSE\\
        \midrule
        Random Forest &0.5644&0.6891 &0.9279 \\
        EGAT &0.6549 &0.5980 &0.8265 \\
        EGNN &0.5800 &0.6662 &0.9111 \\
        \textbf{HENCE} &\textbf{0.7452}&\textbf{0.5188}&\textbf{0.7102} \\
        \bottomrule
    \end{tabular}
    \centering
    \caption{Temporal transfer study. Models are trained on the Year 2015 dataset and tested on the Year 2017 dataset.}
    \label{tbl:temporal}
\end{table}

Road networks and OD are also evolving over time under local development or policy interventions. Accurately estimating future on-road carbon emissions with evolving road networks and OD flows holds great significance for efficiently managing on-road carbon emission evolutions and evaluating carbon-reduction policies. Therefore, we evaluate our model's performance under temporal transfer, training our model on the Year 2015 dataset and applying the trained model to predict on-road carbon emissions in 2017. We also conduct temporal transfer experiments on the three best baselines for comparison. The performance results are listed in Table~\ref{tbl:temporal}. As shown in the table, our model achieves an excellent transfer performance with $R^2$ exceeding 0.74. The performance gap between our model and the best baseline achieved a remarkable value of 13.79\% on the $R^2$ metric. Compared to the performance of being trained and evaluated both on Year 2017 dataset, our model's $R^2$ performance is only slightly lower with a 5.25\% gap. The excellent temporal transferability results indicate our model's capacity to capture the time-invariant intrinsic correlation between OD network, road network, and on-road carbon emissions.

\subsection{Explanatory Insights}
The heterogeneous graphs we employ at both the region level and community level are aimed at modeling the intricate interactions between road networks and OD at different spatial scales. The adaptive attention weights for the links indicate the varying explainability of the two factors for different regions. Therefore, to investigate the correlation between transportation demand, road network connectivity, and emission level, we visualize the aggregation attention weights derived by Equ~\ref{equ:attention} for both types of links in the community-level graph and the region-level graph.

\begin{figure}[h]
    \centering
    \includegraphics[width=0.75\columnwidth]{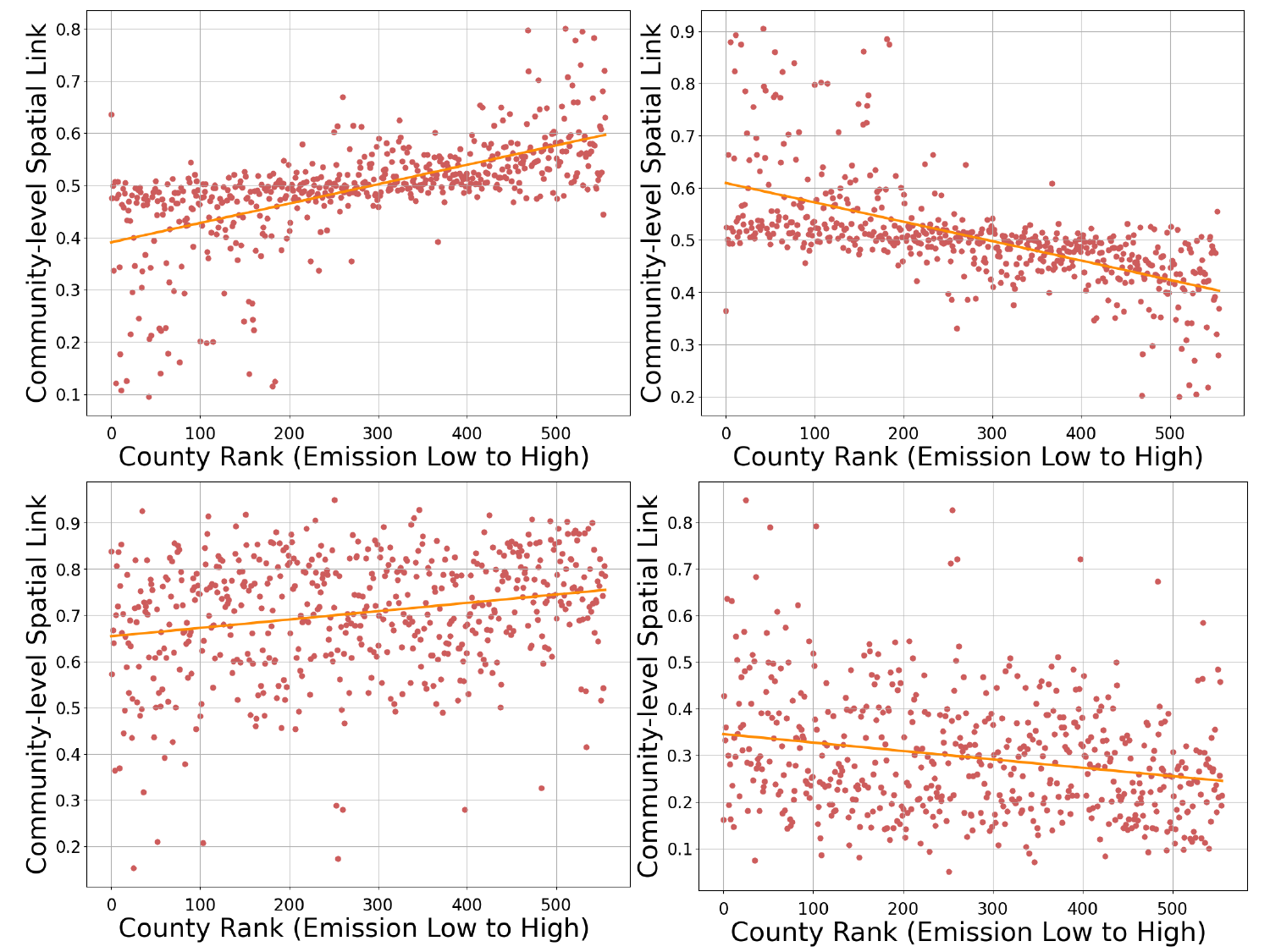}
    \caption{Message aggregation weights for OD links and spatial links of the region level and community level.}
    \label{fig:weights}
\end{figure}

As shown in Figure~\ref{fig:weights}, our model puts more attention on the OD links for regions with larger emissions, whether for community-level short-distance travel or region-level long-distance travel. It may be due to better road construction for large-emission regions where people rely more heavily on road transportation. Therefore road network connectivity is consistently high in these regions and spatial links convey less information. Meanwhile, for regions with relatively low emissions, the road network connectivity significantly impacts their traffic efficiency, accounting for the emission discrepancy for regions with similar transportation demands. Hence, for large-emission regions which are our carbon mitigation priorities, the most efficient carbon mitigation strategy may be better land use planning to decrease travel distances and navigation towards public transportation.

%% file: 7.conclusion.tex
\section{Conclusion}
In this work, we propose an effective hierarchical heterogeneous graph learning method for on-road transportation carbon emission estimation with the open data of road network and origin-destination flow data. With the hierarchical graph modeling multi-scale relations between urban spaces and the heterogeneous graph characterizing the intrinsic interactions between travel demand and road network-based connectivity, HENCE comprehensively models how diverse travel demands are fulfilled by road network-based spatial connectivity. Extensive experiments demonstrate HENCE's effectiveness in complex and diverse application scenarios, proving the great potential of employing artificial intelligence to empower sustainability development.

\section{Acknowledgements}
This work was supported by the National Key Research and Development Program of China under 2022YFF0606904 and the National Natural Science Foundation of China under U20B2060, U21B2036, and 62171260.